\pdfoutput=1

\documentclass[11pt]{article}

\usepackage[]{acl}

\usepackage{times}
\usepackage{latexsym}

\renewcommand{\paragraph}[1]{\vspace{0.2cm}\noindent\textbf{#1}}

\newcommand{\la}{$_\texttt{large}$\xspace}
\newcommand{\ba}{$_\texttt{base}$\xspace}

\newcommand\tf[1]{\textbf{#1}}
\newcommand\uf[1]{\underline{#1}}
\newcommand\ttt[1]{\texttt{#1}}

\newcommand\mr[1]{\mathrm{#1}}

\newcommand{\tableindent}{~~}
\newcommand\footnoteref[1]{\protected@xdef\@thefnmark{\ref{#1}}\@footnotemark}

\newcommand{\cls}{\ttt{[CLS]}}

\usepackage[T1]{fontenc}

\usepackage[utf8]{inputenc}

\usepackage{microtype}

\usepackage{graphicx, multirow}
\usepackage{caption, subcaption}
\usepackage{array}
\usepackage{pifont}
\usepackage{tabularx}
\usepackage{adjustbox}
\usepackage{multirow}
\usepackage{enumitem}
\usepackage{xspace}
\usepackage{tcolorbox}
\usepackage{booktabs,amsfonts,dcolumn}
\usepackage{hyperref}
\usepackage{url}
\usepackage{amsmath,amsthm,amsfonts,amssymb,bm,stmaryrd,bbm}
\usepackage[noorphans,vskip=0.75ex,leftmargin=2ex]{quoting}
\usepackage{booktabs}
\usepackage{float}

\title{Composition-contrastive Learning for Sentence Embeddings}

\author{Sachin Chanchani \and Ruihong Huang \\
  Texas A\&M University \\
  \texttt{\{chanchanis, huangrh\}@tamu.edu}}

\begin{document}
\maketitle
\begin{abstract}

Vector representations of natural language are ubiquitous in search applications. Recently, various methods based on contrastive learning have been proposed to learn textual representations from unlabelled data; by maximizing alignment between minimally-perturbed embeddings of the same text, and encouraging a uniform distribution of embeddings across a broader corpus. Differently, we propose maximizing alignment between texts and a composition of their phrasal constituents. We consider several realizations of this objective and elaborate the impact on representations in each case. Experimental results on semantic textual similarity tasks show improvements over baselines that are comparable with state-of-the-art approaches. Moreover, this work is the first to do so without incurring costs in auxiliary training objectives or additional network parameters.\footnote{Code, pre-trained models, and datasets will be available at \href{https://www.github.com/perceptiveshawty/CompCSE}{github.com/perceptiveshawty/CompCSE}.}

\end{abstract}
\section{Introduction}


Significant progress has been made on the task of learning universal sentence representations that can be used for a variety of natural language processing tasks without task-specific fine-tuning (\citealp{conneau-etal-2017-supervised-infersent}, \citealp{cer-etal-2018-universal}, \citealp{kiros2015skip-thought}, \citealp{logeswaran2018an-quick-thought}, \citealp{giorgi2020declutr}, \citealp{yan2021consert}, 
\citealp{gao-etal-2021-simcse}, \citealp{chuang-etal-2022-diffcse}). Recent works have shown the potential to learn good sentence embeddings without labeled data by fine-tuning pre-trained language models (PLMs) using the unsupervised framework introduced in SimCLR \citep{chen2020simple}, adapted to the natural language processing (NLP) domain. In computer vision (CV), SimCLR exploits a series of transformations (blurs, crops, color distortions, etc.) to construct positive pairs from otherwise unique data points. A cross entropy objective (InfoNCE; \citealp{oord2018nce}) is then applied to minimize distance between representations originating from the same datum, while maximizing the distance to all other points in a mini-batch. The success of the framework in computer vision is due largely to the diversity of augmentations used for creating positive pairs, which leave the identity of the original example intact while reducing pairwise mutual information in the input space (\citealp{tianWhatMakesGood2020b}; \citealp{wuMutualInformationContrastive2020}; \citealp{purushwalkam_demystifying_2020}). 

Constructing positive pairs via discrete augmentations have not been effective when applying the same objective to sentence embeddings. In fact, \citet{gao-etal-2021-simcse} perform an ablation study of textual augmentations (e.g., cropping, synonym replacement) and find that training on these pairs hurts downstream performance on semantic textual similarity (STS) tasks. Instead, they observe that minimal (10\%) dropout noise can be used to create positive pairs on-the-fly, and empirically results in stronger representations. This framework relying on nearly identical pairs is known as SimCSE. Since the dropout noise exists as a regularization component of the BERT architecture \cite{devlin-etal-2019-bert}, explicit augmentations are unnecessary, making it a simple yet effective framework for unsupervised learning of sentence embeddings. 

Here, we make a case for composition as augmentation, by exploiting its presence in language as a signal for learning sentence encoders. We conduct a series of experiments to illustrate the impact of training on positive examples derived by averaging representations of textual constituents in the latent space. Following previous works, we benchmark the proposed strategy on 7 STS tasks. Our results show that it is feasible to significantly improve upon SimCSE without making expensive architectural modifications or changing the overall training objective. We hope our findings can inspire new avenues of inquiry in text representation learning that draw on long-standing notions in semantics and linguistics.

\begin{figure*}[th!]
    \vspace{-0.5cm}
    \centering
    \begin{subfigure}[t]{0.45\textwidth}
        \includegraphics[width=\textwidth, trim=0.5cm 0cm 0.5cm 0cm]{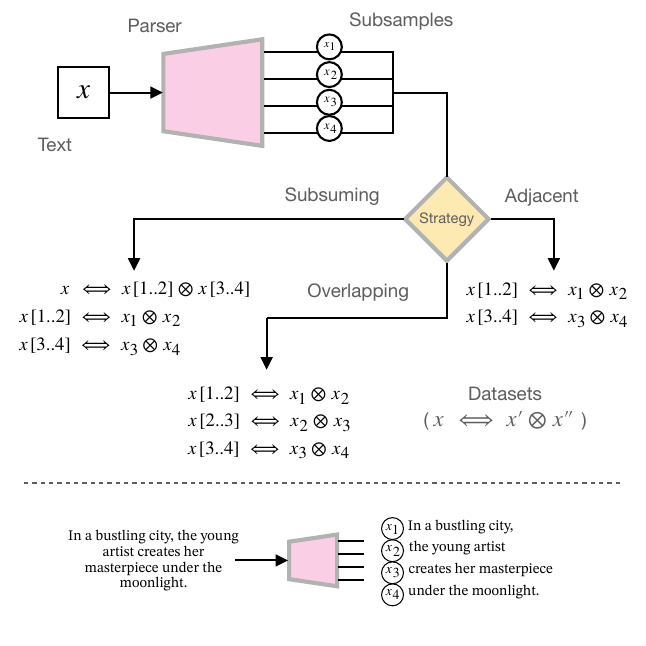}
        \vspace{-0.8cm}
        \caption{
            subsampling strategies
        }
        \label{fig:subsampling}
    \end{subfigure}
    \begin{subfigure}[t]{0.45\textwidth}
        \hspace{2mm}
        \includegraphics[width=\textwidth, trim=1cm 0cm 0cm 0cm, clip]{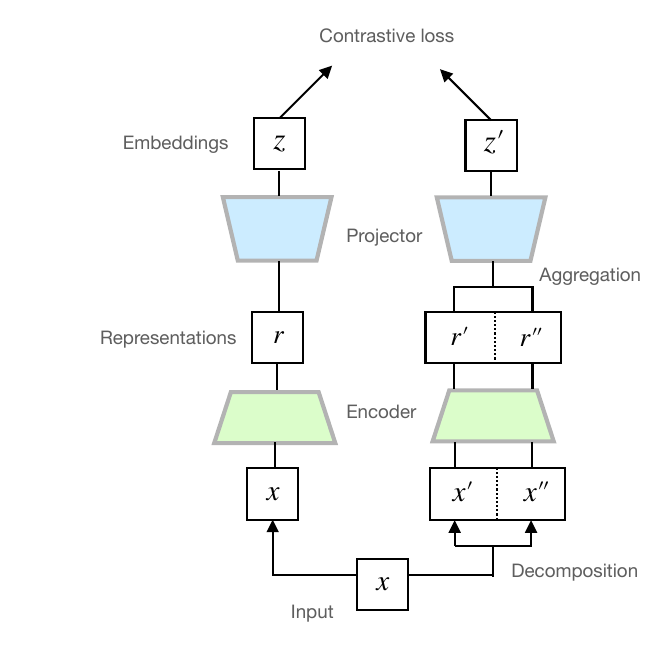}
        \caption{
            training objective
        }
        \label{fig:framework}
    \end{subfigure}%
    \caption{An overview of composition-based contrastive learning. Subsampling strategies used to expand the training dataset 
    are illustrated on the left-hand side, where the bidirectional arrows indicate positive pairs, brackets indicate spans of text, and all other pairs are the standard in-batch negatives. The $\otimes$ operation is a shorthand for the augmentation strategy integrated with the framework and depicted on the right: along with dropout noise, examples are decomposed in the input space and constituents are independently passed through the encoder. Resultant \cls tokens are then aggegrated and passed through a linear projector before computing the contrastive loss.}
    \label{fig:overview}
\end{figure*}

\section{Background and Related Work}

\subsection{Unsupervised Contrastive Learning}
\label{sec:contrastive_learning}

Contrastive learning \cite{hadsell2006dimensionality} aims to learn vector-valued representations of data without relying on annotations. Meaning is derived from these representations based on their proximity to other points in the same space, e.g. two images of dogs will be closer in space than a dog and a chair. Several works have theoretically verified the utility of representations derived from contrastive learning \cite{arora2019latent, Lee2020PredictingWY, Tosh2020ContrastiveLM} under various assumptions; \citet{chen2020simple} showed that SimCLR can even outperform supervised counterparts on CV transfer learning benchmarks. In SimCLR (and SimCSE), the learning objective for an example is:
\begin{equation}\label{eq:infonce}
    l_i = -log \frac{e^{sim(z_i, z^+_i) / \tau}}{\sum^{N}_{j=1}e^{sim(z_i, z^+_j)/\tau}},
\end{equation} 
where $z_i = f(x_i), z_i^+ = f(x_i^+)$ are vector representations of an input and its corresponding augmented positive, $\tau$ is a temperature hyperparameter, $sim(.,.)$ is cosine similarity, and $N$ is batch size. 

\paragraph{Drawbacks of InfoNCE.} In examination of eq. \ref{eq:infonce}, it is evident that InfoNCE uniformly repels examples in the mini-batch besides the minimally augmented positive. Consequentially, the resulting embeddings show poor group-wise discrimination, especially in language, since it is likely that different examples in the batch can have different relative similarities to a given anchor. Another consequence of the unsupervised InfoNCE objective is dimensional collapse, wherein embedding vectors are mostly differentiated by a small proportion of the feature axes; thus under-utilizing the full expressive capacity of the encoder. This was theoretically posited in \citet{jing2022understanding}. They prove that minimal augmentation, coupled with an over-parameterized network, results in low rank solutions to the unsupervised contrastive objective. We hypothesize that this is closely tied to short-cut learning \cite{robinson_can_2021} ---- in the context of sentence embeddings, \citet{wu_esimcse_2022} observed that spurious features related to the lengths of sentences are relied on to solve the contrastive objective. Such solutions can yield non-generalizable features that poorly represent data from new domains.

\paragraph{Qualifying the representation space.} \citet{wang2020understanding} proposed two metrics to measure the quality of embeddings derived through contrastive learning. First, \textit{alignment} measures on average the proximity of pairs of examples that \textit{should} be close in space, i.e. for a set of positive pairs $p_{pos}$ and their normalized representations $f(x), f(x^+)$:
\begin{equation}
    \label{eq:alignment}
    \resizebox{.73\hsize}{!}{%
    $
    \ell_{\mr{align}}\triangleq \underset{(x, x^+)\sim p_{\mr{pos}}}{\mathbb{E}} \Vert f(x) - f(x^+) \Vert^2.
    $
    }
\end{equation}
Conversely, \textit{uniformity} measures how scattered the embeddings are upon the unit hypersphere:
\begin{equation}
    \resizebox{.85\hsize}{!}{%
    $
    \label{eq:uniformity}
    \ell_{\mr{uniform}}\triangleq\log \underset{~~~x, y\stackrel{i.i.d.}{\sim} p_{\mr{data}}}{\mathbb{E}}   e^{-2\Vert f(x)-f(y) \Vert^2},
    $
    }
\end{equation}
where $p_{data}$ denotes the full data distribution. We use these metrics to explore the advantages and drawbacks of various augmentations in contrastive pre-training, similarly to \citet{gao-etal-2021-simcse}. 

\subsection{Learning Sentence Embeddings}
\label{sec:sentence_embeddings}

\paragraph{Early works.} First approaches to learning sentence embeddings span unsupervised 
\cite{kiros2015skip-thought,hill-etal-2016-learning,logeswaran2018an-quick-thought}, and supervised \cite{conneau-etal-2017-supervised-infersent, cer-etal-2018-universal, reimers-gurevych-2019-sentence} methods which have been studied extensively in the literature. More recent work has focused on unsupervised contrastive learning with the advent of SimCSE \cite{gao-etal-2021-simcse}, which passes the same sentence to a language model twice; the independent dropout masks sampled in the two forward passes encode the sentence at slightly different positions in vector space. A cross-entropy objective is then used to maximize the probability of top-1 proximity between positives while uniformly repelling other examples. 

\paragraph{Successors to SimCSE.} Works that follow SimCSE attempt to improve the framework with auxiliary training objectives \cite{chuang-etal-2022-diffcse, nishikawa-etal-2022-ease, zhou2023learning, zhang-etal-2022-contrastive, wu_infocse_2022, wang_sncse_2022}, verbalized or continuous prompts \cite{wang_sncse_2022, jiang2022promcse}, instance generation or weighting strategies \cite{zhou-etal-2022-debiased}, momentum encoders with negative sample queues \cite{he_momentum_2020}, or entirely new parameters with secondary networks \cite{wu_pcl_2022}. Many works combine several of these components, making it difficult to discern their impact in isolation. As the design choices have become more intricate and less parameter-efficient, performance on STS benchmarks has too become saturated.  

\section{Composition-based Contrastive Learning}
\label{sec:method}

Our augmentation strategy retains the simplicity and efficiency of SimCSE, as illustrated in Figure \ref{fig:overview}. Specifically, it requires just one additional forward pass that is ultimately compensated by a non-trivial reduction in convergence time (\S\ref{sec:analysis}). Beginning with a corpus of unlabelled sentences $\{x_i\}_{i=1}^m$,  we consider $x_i^+$ only in the latent space, as a composition of the representations of $(x_i^{'+}, x_i^{''+})$. A simple (and effective) way to curate $(x_i^{'+}, x_i^{''+})$ is to split the tokens of $x_i$ in half, and encode the left and right phrases in independent forward passes through the encoder and linear projector. After obtaining their respective \cls token representations $(z_i, z_i^{'+}, z_i^{''+})$, $(z_i^{'+}, z_i^{''+})$ is aggregrated and taken to be the corresponding positive example for $z_i$. The training objective for a single pair is then the same as in eq.\xspace\ref{eq:infonce}, where $z^+ = aggregate(z_i^{'+}, z_i^{''+}).$ We experiment with aggregation methods in \S\ref{sec:ablation}, and find that the best approach varies according to the size and type of underlying PLM. In our final model based on BERT\ba, we find that this manner of augmentation is especially suitable for the scheme proposed in DirectCLR \cite{jing2022understanding}, which aims to directly mitigate dimensional collapse by computing the loss from eq.\xspace \ref{eq:infonce} on a subset of the embedding vector axes before backpropagating to the entire representation. 

\paragraph{Decomposition as data augmentation.} To explain the motivation for decomposing examples in the input space, we can consider an example from the development subset of STS-B labelled as having high semantic similarity:

\begin{small}
\begin{verbatim}
A man is lifting weights in a garage.
A man is lifting weights.
\end{verbatim}
\end{small}

There are two semantic atoms at play in the first text: 1) a man is lifting weights, and 2) a man is in a garage. The similarity between the two texts can only be considered high based on the first atom; lifting weights. It cannot be said that there is a general relation between \textit{being in a garage} and \textit{lifting weights} - a garage is equally, if not more likely to be related to cars, parking, or storage, yet this does not preclude a connection between them. It is only through the composition of both atoms that we can relate the two. Thus, there is a need for sentence encoders to learn more generalized phrase representations; to at least implicitly abide by principles of semantic compositionality. The challenge in enforcing this kind of constraint through a contrastive objective is in the choice of data --- it would require a corpus where lexical collocations are encountered across a diverse set of contexts. 
\begin{figure}[th!]
    \centering
    \includegraphics[width=\columnwidth, trim= 0cm 0cm 0cm 0cm]{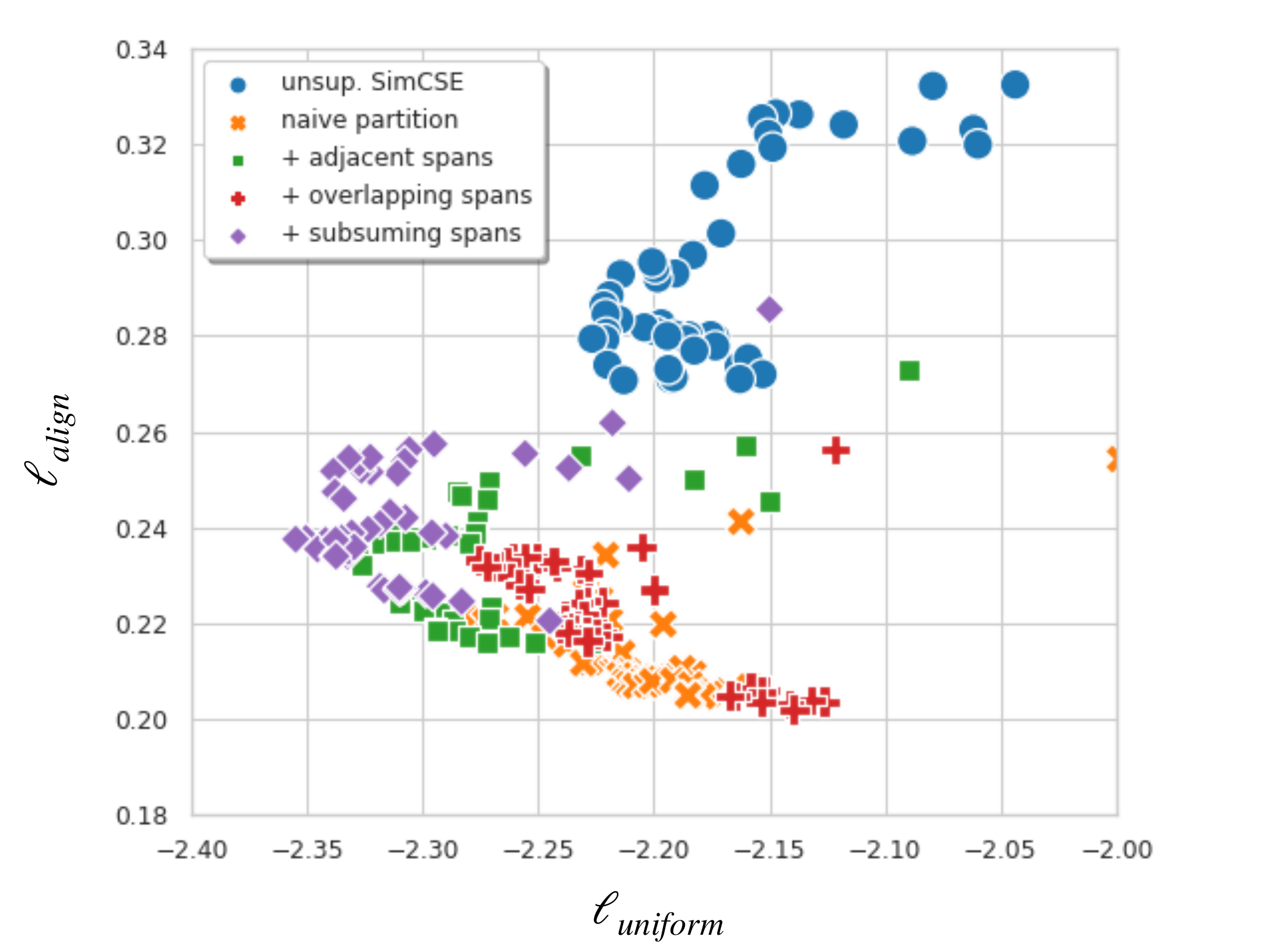}
    \caption{$\ell_{\mr{align}}$-$\ell_{\mr{uniform}}$ tradeoff for subsampling strategies explored in this work. Measurements are taken every 10 training steps on the development subset of STS-B, for 500 steps with BERT\ba. The ideal trajectory is the negative direction for both axes/metrics.}
    \label{fig:align_uniform_base}

\end{figure}

\paragraph{Subsampling from decomposed inputs.} To further examine the effect of decomposition in the input space, we leverage a pre-trained discourse parser\footnote{https://github.com/seq-to-mind/DMRST\_Parser} to extract atomic semantic units from each unique example in the training set; typically simple phrases or clauses. We experiment with 3 kinds of strategies (Figure \ref{fig:subsampling}) to expand the training set, besides considering our augmentation in isolation: let $C = \{x_{i, k}\}_{k=1}^c$ represent the $c$ non-overlapping phrases extracted from an input $x_i:$ 
\begin{itemize}
    \item \textbf{adjacent spans} are sampled by taking each unique pair in $C$ such that there is no overlap between inputs;
    \item \textbf{overlapping and adjacent spans} are sampled by taking (potentially) overlapping pairs in $C$;
    \item \textbf{overlapping, adjacent, and subsuming spans} are sampled by recursively partitioning the elements of $C$ in half, i.e. maximizing the lexical overlap of extracted input samples.  
\end{itemize}

\paragraph{Impact on the representation space.} A consequence of expanding the training set with subsamples is the presence of harder in-batch negatives. Prior work has demonstrated that this is generally beneficial to contrastive learning \cite{robinson2021contrastive, kalantidis2020hard, zhang-stratos-2021-understanding}. Following \citet{gao-etal-2021-simcse}, we measure the uniformity and alignment of representations obtained for the development set of STS-B to understand the effect of training with additional subsamples. STS-B is comprised of pairs of sentences accompanied by a score between 1-5 indicating degree of semantic similarity. We take all pairs as $p_{\mr{data}},$ and pairs with a score greater than 4 as $p_{\mr{pos}}.$ Both metrics are measured every 10 steps for 500 training steps, to understand the direction in which each of our strategies drives the encoder. 

\bigskip

As shown in Figure \ref{fig:align_uniform_base}, any of the subsampling strategies can bring non-trivial improvements over unsupervised SimCSE in both alignment and uniformity. Specifically, expanding the training set with subsamples (\textit{+ adjacent, + overlapping, + subsuming}) encourages a more uniform embedding distribution. On the other hand, forgoing subsampling for just the compositional augmentation (\textit{naive partition}) achieves the better alignment while retaining the uniformity of SimCSE. This is because we leave the self-prediction objective intact, while increasing its difficulty: although subsamples are potentially highly related, positive pairs are only curated from the exact same text. As a consequence, the underlying PLM is forced to effectively distinguish examples with high lexical overlap --- which is precisely the intuition underlying DiffCSE \citet{chuang_diffcse_2022}, and other discriminative pre-training objectives. 


\section{Experiment}
\label{sec:experiment}

\begin{table*}[th!]
    \begin{center}
    \footnotesize
    \centering
    \begin{tabular}{clcccccccc}
    \toprule
       \tf{PLM} & \tf{Method} & \tf{STS12} & \tf{STS13} & \tf{STS14} & \tf{STS15} & \tf{STS16} & \tf{STS-B} & \tf{SICK-R} & \tf{Avg.} \\
    \midrule
        \multirow{9}{5em}{BERT\ba} & SimCSE$\clubsuit$ & 68.40 & 82.41 & 74.38 & 80.91 & 78.56 & 76.85 & 72.23 & 76.25 \\ 
                                    & L2P-CSR$\heartsuit$ & 70.21 & 83.25 & 75.42 & 82.34 & 78.75 & 77.8 & 72.65 & 77.20 \\
                                    & DCLR$\spadesuit$ & 70.81 & 83.73 & 75.11 & 82.56 & 78.44 & 78.31 & 71.59 & 77.22 \\
                                    & MoCoSE$\diamondsuit$ & 71.58 & 81.40 & 74.47 & \uf{83.45} & 78.99 & 78.68 & 72.44 & 77.27 \\
                                    & ArcCSE\textdagger & 72.08 & \uf{84.27} & 76.25 & 82.32 & 79.54 & 79.92 & 72.39 & 78.11 \\
                                    & PCL\textdaggerdbl & \uf{72.74} & 83.36 & 76.05 & 83.07 & 79.26 & 79.72 & \uf{72.75} & 78.14 \\
                                    & $*$SimCSE (\textit{w/ comp.}) & 72.14 & 84.06 & 75.38 & \bf 83.82 & \uf{80.43} & \uf{80.29} & 71.12 & 78.18 \\
                                    & ESimCSE\textbigcircle & \bf 73.40 & 83.27 & \bf 77.25 & 82.66 & 78.81 & 80.17 & 72.30 & \uf{78.27} \\
                                    & SNCSE\textdollar & 70.67 & \bf 84.79	& \uf{76.99} & \uf{83.69} & \bf 80.51 & \bf 81.35 & \bf 74.77 & \bf 78.97 \\

        \midrule
        \multirow{9}{5em}{BERT\la} & SimCSE $\clubsuit$ & 70.88 & 84.16 & 76.43 & 84.50 & 79.76 & 79.26 & 73.88 & 78.41 \\ 
                                   & DCLR $\spadesuit$ & 71.87 & 84.83 & 77.37 & 84.70 & 79.81 & 79.55 & 74.19 & 78.90 \\
                                   & L2P-CSR $\heartsuit$ & 71.44 & 85.09 & 76.88 & 84.71 & 80.00 &	79.75 & 74.55 & 78.92 \\
                                   & MoCoSE $\diamondsuit$ & 74.50 & 84.54 & 77.32 & 84.11 & 79.67 & 80.53 & 73.26 & 79.13 \\
                                   & ESimCSE\textbigcircle & 73.21 & 85.37 & 77.73 & 84.30 & 78.92 & 80.73 & \uf{74.89} & 79.31 \\
                                   & ArcCSE\textdagger & 73.17 & 86.19 & 77.90 & 84.97 & 79.43 & 80.45 & 73.50 & 79.37 \\
                                   & $*$SimCSE (\textit{+ subsum.}) & \bf 75.10 & \uf{86.57} & 77.70 & 84.72 & \uf{80.25} & 80.17 & 73.21 & 79.67 \\
                                   & PCL\textdaggerdbl & \uf{74.89} & 85.88 & \uf{78.33} & \uf{85.30} & 80.13 & \uf{81.39} & 73.66 & \uf{79.94} \\
                                   & SNCSE\textdollar	& 71.94 & \bf 86.66 & \bf 78.84 & \bf 85.74 & \bf 80.72 & \bf 82.29 & \bf 75.11 & \bf 80.19 \\

        \midrule
        \multirow{7}{5em}{RoBERTa\ba} & SimCSE$\clubsuit$                 & 70.16 & 81.77 & 73.24 & 81.36 & 80.65 & 80.22 & 68.56 & 76.57 \\ 
                                      & ESimCSE\textbigcircle             & 69.90 & 82.50 & 74.68 & 83.19 & 80.30 & 80.99 & \uf{70.54} & 77.44 \\
                                      & L2P-CSR$\heartsuit$               & \uf{71.69} & 82.43 & 74.55 & 82.15 & \bf 81.81 & 81.36 & 70.22 & 77.74 \\
                                      & DCLR$\spadesuit$                  & 70.01 & 83.08 & 75.09 & \uf{83.66} & 81.06 & \uf{81.86} & 70.33 & 77.87 \\
                                      & $*$SimCSE (\textit{w/ comp.})     & \bf 72.56 & \uf{83.33} & 73.67 & 83.36 & 81.14 & 80.71 & 70.39 & 77.88 \\
                                      & PCL\textdaggerdbl                 & 71.54 & 82.70 & \uf{75.38} & 83.31 & \uf{81.64} & 81.61 & 69.19 & \uf{77.91} \\
                                      & SNCSE\textdollar                  & 70.62 & \bf 84.42	& \bf 77.24 & \bf 84.85 & 81.49 & \bf 83.07 & \bf 72.92 & \bf 79.23 \\
        \midrule
        \multirow{7}{5em}{RoBERTa\la} & $*$SimCSE (\textit{w/ comp.}) & 72.32 & 84.19 & 75.00 & 84.83 & 81.27 & 82.10 & 70.99 & 78.67 \\ 
                                    & SimCSE$\clubsuit$ & 72.86 & 83.99 & 75.62 & 84.77 & 81.80 & 81.98 & 71.26 & 78.90 \\
                                    & DCLR$\spadesuit$ & 73.09 & 84.57 & 76.13 & 85.15 & 81.99 & 82.35 & 71.80 & 79.30 \\
                                    & PCL\textdaggerdbl & \bf 73.76 & 84.59 & 76.81 & 85.37 & 81.66 & \uf{82.89} & 70.33 & 79.34 \\
                                    & ESimCSE\textbigcircle & 73.20 & \uf{84.93} & \uf{76.88} & 84.86 & 81.21 & 82.79 & 72.27 & 79.45 \\
                                    & L2P-CSR$\heartsuit$ & 73.29 & 84.08 & 76.65 & \uf{85.47} & \uf{82.70} & 82.15 & \uf{72.36} & \uf{79.53} \\
                                    & SNCSE\textdollar & \uf{73.71} & \bf 86.73 & \bf 80.35 & \bf 86.80 & \bf 83.06 & \bf 84.31 & \bf 77.43 & \bf 81.77 \\

    \bottomrule
    \end{tabular}
    \end{center}
    \caption{
        The performance on STS tasks (Spearman's correlation) for different sentence embedding models. Results are imported as follows --- $\clubsuit$: \citet{gao-etal-2021-simcse}, $\heartsuit$: \citet{zhou2023learning}, $\spadesuit$: \citet{zhou-etal-2022-debiased}, $\diamondsuit$: \citet{cao-etal-2022-exploring}, \textdagger: \citet{zhang-etal-2022-contrastive}, \textdaggerdbl: \citet{wu_pcl_2022}, \textbigcircle: \citet{wu_esimcse_2022}, \textdollar: \citet{wang_sncse_2022}, $*$: our results.
    }
    \label{tab:main_sts}
\end{table*}

\paragraph{Setup.} In our experiments, we modify the public PyTorch implementation\footnote{https://github.com/princeton-nlp/SimCSE} of SimCSE to support our proposed augmentation and subsampling methods. All of our language models are initialized from pre-trained BERT/RoBERTa checkpoints \cite{devlin_bert_2019, liu2019roberta}, except the randomly-initialized MLP over the \cls representation. For all models, we employ the scheme illustrated in Figure \ref{fig:overview} and report the best results after training with or without the 3 subsampling strategies. We keep the best checkpoints after evaluating on the development set of STS-B every 125 steps during training. Batch size is fixed at 64 for all models; for base and large sized models, learning rates are fixed to $3\text{e-}5$ and $1\text{e-}5$ respectively. Besides those covered in \ref{sec:ablation}, extensive hyperparameter searches were not conducted in this work.

\paragraph{Data.} We use the same 1 million randomly sampled sentences\footnote{https://huggingface.co/datasets/princeton-nlp/datasets-for-simcse} as SimCSE for training, besides incorporating the subsampling strategies from \S\ref{sec:method}. We evaluate on 7 semantic textual similarity tasks: STS 2012-2016, STS-Benchmark, SICK-Relatedness \cite{agirre-etal-2012-semeval,agirre-etal-2013-sem,agirre-etal-2014-semeval,agirre-etal-2015-semeval,agirre-etal-2016-semeval,cer-etal-2017-semeval,marelli-etal-2014-sick} and report averaged Spearman's correlation across all available test subsets. We employ the modified SentEval\footnote{https://github.com/facebookresearch/SentEval} \cite{conneau_senteval_2018} package accompanying the source code of SimCSE for fair comparison with other works. 

\paragraph{Baselines.} We compare our results with many contemporaries: ESimCSE \cite{wu_esimcse_2022}, SNCSE \cite{wang_sncse_2022}, PCL \cite{wu_pcl_2022}, DCLR \cite{zhou-etal-2022-debiased}, ArcCSE \cite{zhang-etal-2022-contrastive}, MoCoSE \cite{cao-etal-2022-exploring}, and L2P-CSR \cite{zhou2023learning}. We consider SimCSE \cite{gao-etal-2021-simcse} as our baseline, since we leave its training objective and network architecture intact. 

\paragraph{Results.} We can observe in Table \ref{tab:main_sts} that our methods bring non-trivial improvements to SimCSE with both BERT encoders, as well as RoBERTa\ba. In fact, we achieve an average F1 score within 0.8 points of SNCSE-BERT\ba \cite{wang_sncse_2022}. SNCSE exploits biases in test sets by engineering hard negatives via explicitly negated sentences --- the impact of this strategy is more apparent in the results utilizing RoBERTa, where there is parity in all works besides SNCSE. In the case of BERT\la, the gap in performance between our approach and SNCSE is narrower at 0.52 points. A clear failure of the composition-augmented objective presents itself in the results with RoBERTa\la. This could be attributed to poor hyperparameter settings, or a fundamental incompatibility between our approach and the model size/RoBERTa pre-training objective, since other works achieve better results with this PLM.

\section{Ablation}
\label{sec:ablation}

 We ablate several aspects of the approach to understand their impact in isolation. We first consider the subsampling strategy, or lack thereof, in which each model achieves the best STS-B development set performance. These are then tied to each model in subsequent ablations. 

 \paragraph{Including subsamples.} In the process of designing DeCLUTR, \citet{giorgi-etal-2021-declutr} report gains from subsampling more than one anchor per input document. In our experiments, we find that the aligment-uniformity trade-off differs between BERT\la and BERT\ba, ie. different strategies can be better suited to different PLMs. In Table \ref{tab:sampling}, we show that including subsamples is beneficial to the BERT\la PLM, but harmful to BERT\ba. This is likely a result of the difference in no. of parameters --- the smaller PLM may not possess the expressive capacity to distinguish highly related texts without suffering a degeneration in alignment. With RoBERTa\ba, we observe that subsampling non-overlapping spans gives the best results, whereas none of our strategies appeared compatible with RoBERTa\la.
 \begin{table}[th!]
    \begin{center}
    \centering
    \small
    \begin{tabular}{llc}
    \toprule
         \tf{PLM} & \tf{Method} & \tf{STS-B} \\
    \midrule
    \multirow{5}{6em}{BERT\ba} & SimCSE & 81.47 \\
        & ~~ \textit{w/ composition} & \tf{83.97} \\ 
    \cmidrule{2-3}
        & Additional subsampling: & \\
            & ~~\textit{+ adjacent} & 83.39 \\
            & ~~\textit{+ overlapping} & 83.18 \\
            & ~~\textit{+ subsuming} & 82.97 \\
    \midrule
    \multirow{5}{6em}{BERT\la} & SimCSE & 84.41 \\
        & ~~\textit{w/ composition} & 84.79 \\ 
    \cmidrule{2-3}
        & Additional subsampling: & \\
            & ~~\textit{+ adjacent} & 84.84 \\
            & ~~\textit{+ overlapping} & 85.01 \\
            & ~~\textit{+ subsuming} & \tf{85.06} \\
    \midrule
    \multirow{5}{6em}{RoBERTa\ba} & SimCSE & 83.91 \\
        & ~~\textit{w/ composition} & \tf{84.14} \\ 
    \cmidrule{2-3}
        & Additional subsampling: & \\
            & ~~\textit{+ adjacent} & 84.00 \\
            & ~~\textit{+ overlapping} & 84.10 \\
            & ~~\textit{+ subsuming} & 82.92 \\
    \midrule
    \multirow{5}{6em}{RoBERTa\la} & SimCSE & \bf 85.07 \\
        & ~~\textit{w/ composition} & \uf{84.80} \\ 
    \cmidrule{2-3}
        & Additional subsampling: & \\
            & ~~\textit{+ adjacent} & 83.91 \\
            & ~~\textit{+ overlapping} & 82.74 \\
            & ~~\textit{+ subsuming} & 83.33 \\
    \bottomrule
    \end{tabular}
    \end{center}
    \caption{
        Development set results of STS-B after varying the subsampling strategy on different-sized PLMs. 
    }
    \label{tab:sampling}
\end{table}

 \paragraph{Aggregration method.} In SBERT \cite{reimers-gurevych-2019-sentence}, and in historically effective works such as InferSent \cite{conneau-etal-2017-supervised-infersent}, PLMs are fine-tuned with a cross entropy loss to predict whether two sentences $u$ and $v$ entail or contradict eachother. Their pooler is a concatenation of the two sentence embeddings, along with second-order features such as the element-wise difference, $|u - v|.$ We experiment with these aggregration methods, as well as simpler choices such as element-wise sums/averages. We can see in Table \ref{tab:aggregration} that simply interpolating the embeddings is preferable to other methods for BERT-based encoders. We postulate that this interpolation functions as a form of self-distillation, and amplifies the salience of desirable sparse features correlated with sentential context \cite{wenUnderstandingFeatureLearning2021}. For RoBERTa, we find that concatenating the first and last halves of the representations is better. Since RoBERTa does not use the next-sentence prediction (NSP) objective, its embeddings will not encode sentential knowledge. Averaging RoBERTa embeddings may not correlate well with real tokens in its vocabulary, whereas concatenating the first and last halves of constituent embeddings retains localized token-level information, making it a better choice in this case.
 \begin{table}[th!]
    \begin{center}
    \centering
    \small
    \resizebox{0.9\columnwidth}{!}{%
    \begin{tabular}{lcc}
    \toprule
        \tf{Aggregration} & \tf{STS-B} \\
    \midrule
        BERT\ba \\
        \tableindent \tableindent sum & 83.92 \\
        \tableindent \tableindent avg. & \bf 83.97 \\
        \tableindent \tableindent concat first \& last half & 83.01 \\
        \tableindent \tableindent concat + project & 69.24 \\
        \tableindent \tableindent concat w/ abs. difference + project & 68.79 \\
        \midrule
    RoBERTa\ba \\
        \tableindent \tableindent sum & 84.00 \\
        \tableindent \tableindent avg. & 84.08 \\
        \tableindent \tableindent concat first \& last half & \bf 84.14 \\
        \tableindent \tableindent concat + project & 65.02 \\
        \tableindent \tableindent concat w/ abs. difference + project & 65.44 \\
    \bottomrule
    \end{tabular}
    }
    \end{center}

    \caption{
        Results of different aggregration methods for composing $z^+$ in the latent space. Results are based on BERT\ba on the development set of STS-B. 
    }
    \label{tab:aggregration}
\end{table}

\paragraph{Composing $\pmb{z}$ vs. $\pmb{z^+}$.} In our training objective, there are two sets of sentence representations, one derived from pure dropout noise, and the second by averaging the coordinates of constituent representations. However, for each sentence we can: 1) compose the anchor $z$ in latent space, which means other in-batch examples are repelled from a synthetic example's coordinate, 2) compose the positive $z^+$, which means synthetic coordinates are repelled from representations of real examples, or 3) compose both $z$ and $z^+$ in the latent space. In Table \ref{tab:compose-hyper}, we can see that with BERT\ba, we found the best results by directly embedding the anchor sentence, and composing $z^+$ from constituents. 
\begin{table}[h]
    \begin{center}
    \centering
    \small
    \begin{tabular}{lcccc}
    \toprule
        \multirow{2}{5em}{BERT\ba} & \tf{Compose} & \multicolumn{1}{c}{\it ${z}$ } & \multicolumn{1}{c}{\it ${z^+}$ } & \multicolumn{1}{c}{\it Both} \\
        & \bf STS-B & 83.61 & \bf 83.97 & 83.81 \\
    \bottomrule
    \end{tabular}
    \end{center}
    \caption{
        Differences between having compositional anchors and positives. In the \textit{Both} case, the model framework is symmetric in that both anchors and positives are composed of constituent representations. Results are based on BERT\ba on the development set of STS-B.
    }
    \label{tab:compose-hyper}
\end{table}

\paragraph{Number of partitions.} Within our framework, we can aggregrate the embeddings of two or more phrases. Increasing the number of phrases increases the number of forward passes, and magnifies the impact of dropout noise. We find that partitioning into more than two bins is detrimental to the objective (Table \ref{tab:partition-hyper}), though perhaps this is the case because the evaluation data consists mostly of short-length sentences. 
\begin{table}[th!]
    \begin{center}
    \centering
    \small
    \begin{tabular}{lcccc}
    \toprule
        \multirow{2}{5em}{BERT\ba} & \tf{Partitions} & \multicolumn{1}{c}{\it $2$ } & \multicolumn{1}{c}{\it $3$} & \multicolumn{1}{c}{\it $4$} \\
        & \bf STS-B & \bf 83.97 & 83.48 & 83.52 \\
    \bottomrule
    \end{tabular}
    \end{center}
    \caption{
        Impact of splitting examples into more than 2 bins.  Results are based on BERT\ba with the development set of STS-B.
    }
    \label{tab:partition-hyper}
\end{table}

\paragraph{Hyperparameter $\pmb{d_0}$.} In our experiments with BERT\ba, computing the contrastive loss on a subvector of $(z_i, z_i^+)$ is complementary to composing $z_i^+$ in the latent space. When $d_0 \rightarrow d$, our training objective is the exact same as in all *CSE works, ie. computing the loss on all coordinates of $(z_i, z_i^+)$. For BERT\ba, we search $d_0 \in \{192, 256, 384\}$ with the compositional augmentation in isolation (\textit{w/ composition}); for BERT\la, $d_0 \in \{320, 384, 512\}$ with the expanded training set of subsamples (\textit{+ subsuming}). Our results in Table \ref{tab:d0-hyper} indicate that taking a subvector to compute the loss is beneficial for BERT\ba, but the entire vector is necessary for BERT\la. With RoBERTa encoders, we aggregrate embeddings by concatenating the first and last halves of the phrase embeddings, so $d_0$ is inapplicable.
\begin{table}[th!]
    \begin{center}
    \centering
    \small
    \resizebox{0.96\columnwidth}{!}{%
    \begin{tabular}{lccccc}
    \toprule
        \multirow{2}{5em}{BERT\ba} & $\pmb{d_0}$ & \multicolumn{1}{c}{\it 192} & \multicolumn{1}{c}{\it 256} & \multicolumn{1}{c}{\it 384} & \multicolumn{1}{c}{\it 768} \\
        & \bf STS-B & 83.88 & \bf 84.11 & 83.17 & 83.97 \\
    \midrule
        \multirow{2}{5em}{BERT\la} & $\pmb{d_0}$ & \multicolumn{1}{c}{\it 320} & \multicolumn{1}{c}{384} & \multicolumn{1}{c}{\it 512} & \multicolumn{1}{c}{\it 1024} \\
        & \bf STS-B & 84.61 & 84.94 & 84.98 & \bf 85.06 \\
    \bottomrule
    \end{tabular}
    }
    \end{center}
    \caption{
        Varying the size of the subvector used to compute InfoNCE, as proposed in \citet{jing2022understanding}. Results are based on the development set of STS-B.
    }
    \label{tab:d0-hyper}
\end{table}

\section{Analysis}
\label{sec:analysis}

\begin{figure}[th!]
    \centering
    \includegraphics[width=0.91\columnwidth, trim=0mm 2mm 0mm 3mm]{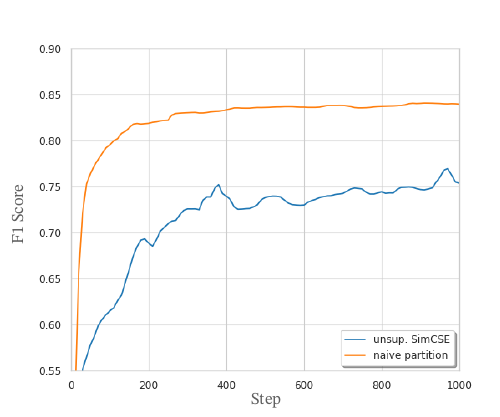}

    \caption{Evaluation curve for BERT\ba, using dropout noise as augmentation (unsup. SimCSE) and latent space composition (naive partition). The y-axis reflects performance on the development set of STS-B.}
    \label{fig:sts_curve}
\end{figure}

\paragraph{Stability and efficiency of training.} Successors to SimCSE have incrementally improved STS performance while disproportionately driving up resource requirements. This limits accessibility to practitioners who wish to learn embeddings from their own corpora, perhaps in other languages.  Differently, our approach relies on a single additional forward pass while converging much faster than SimCSE. In Figure \ref{fig:sts_curve}, we compare our BERT\ba model's evaluation curve to SimCSE's for 1000 training steps in the same setting. We observe that composition as augmentation greatly speeds up convergence, with evaluation metrics plateauing much faster, and more stably than SimCSE. In fact, on a single NVIDIA A100 GPU (40GB), our model can finish training in under 15 minutes.

\begin{figure*}[th!]
    \centering
    \includegraphics[width=0.7\textwidth, trim=10mm 2mm 10mm 2mm]{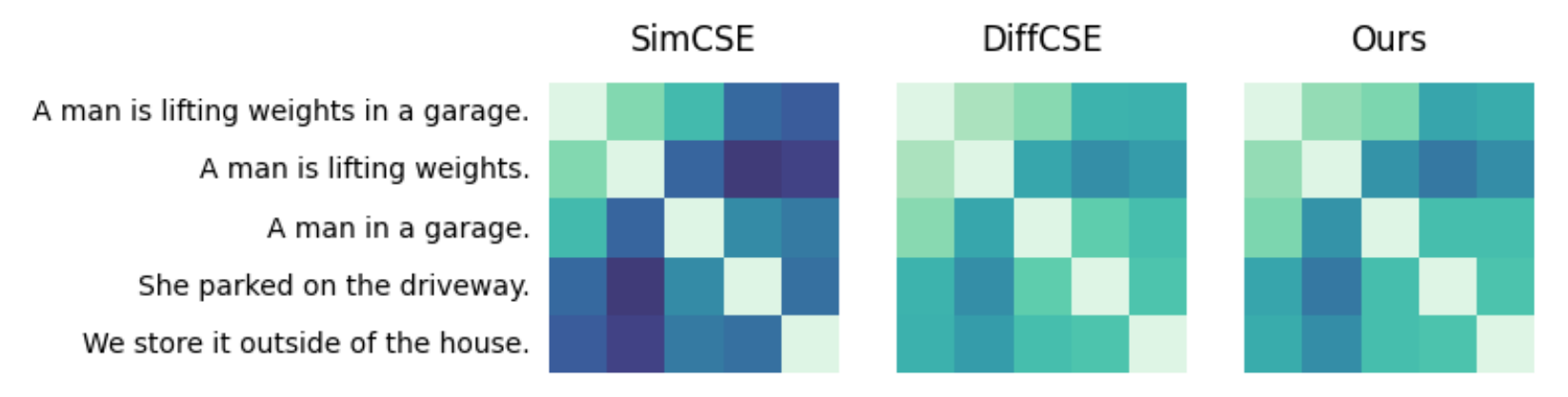}
    \caption{A heatmap displaying the pairwise similarities computed by SimCSE, DiffCSE, and our model, with the same underlying PLM BERT\ba and color scale. Lighter colors indicate higher similarity.}
    \label{fig:retrieval}
\end{figure*}

\begin{figure}[h]
    \centering
    \begin{subfigure}[b]{0.72\columnwidth}
    \includegraphics[width=\columnwidth, trim= 0cm 0cm 0cm 0cm]{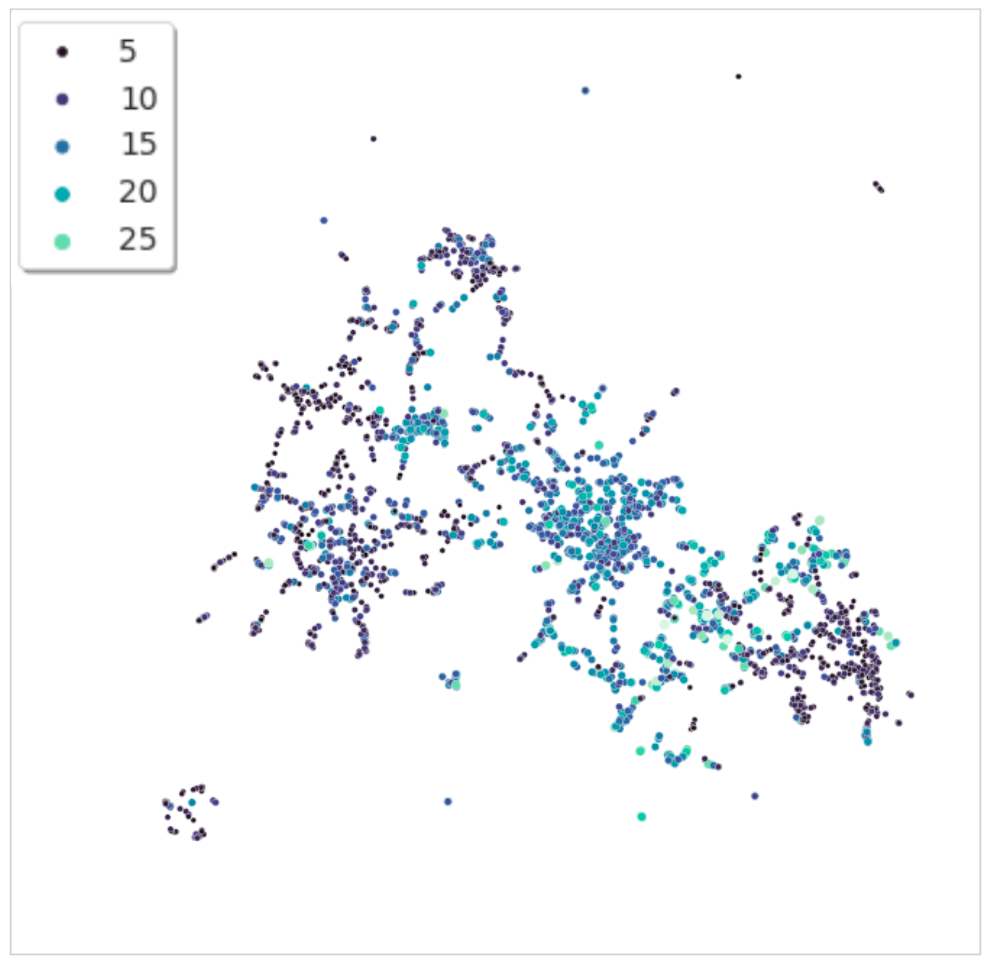}
    \caption{unsup. SimCSE}
    \vspace{2mm}
    \label{fig:simcse_embedding_space}
    \end{subfigure}
    \begin{subfigure}[b]{0.72\columnwidth}
    \includegraphics[width=\columnwidth, trim= 0cm 0cm 0cm 0cm]{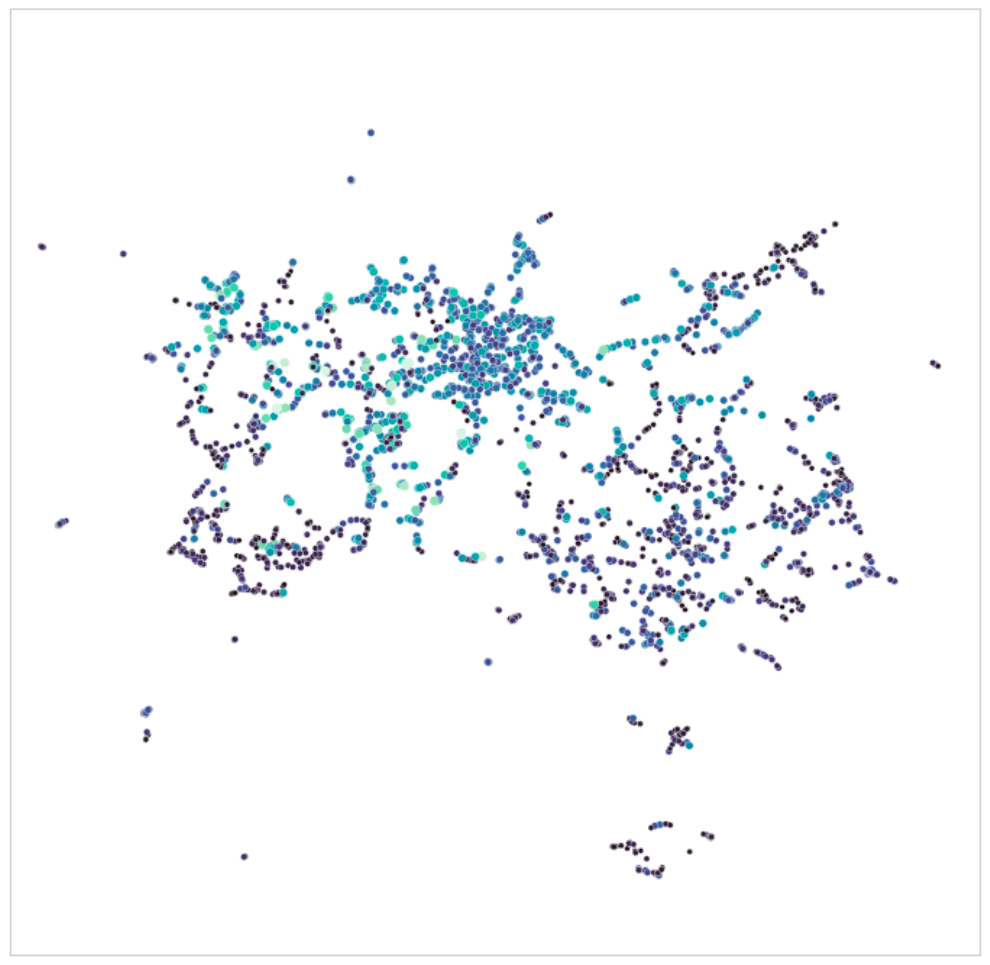}
    \caption{w/ compositional augmentations}
    \label{fig:our_embedding_space}
    \end{subfigure}
    \caption{2D UMAP projection of the representations of all sentences from the validation subset of STS-B. Color indicates word count.
    }
    \label{fig:embedding-space}
\end{figure}
\paragraph{Text length as a feature.} To investigate the structure of the learned space, In Figure \ref{fig:embedding-space}, we visualize embeddings of sentences from the development set of STS-B after down-projecting to 2D Euclidean space. We employ UMAP \cite{mcinnes2018UMAP} with cosine distance as the metric to preserve local and global topological neighborhoods. The same parameters are used to compute the embeddings in Figure \ref{fig:simcse_embedding_space} and \ref{fig:our_embedding_space}, which are derived from dropout noise, and composition-based augmentations (\textit{w/ composition}) respectively. In Figure \ref{fig:simcse_embedding_space}, we can observe several clusters of dark points corresponding to shorter sentences. This corroborates our intuition that minimal augmentation to create positive pairs can lead to shortcut learning, wherein text length is relied upon to solve the training objective. In contrast, we see a more scattered distribution of points in Figure \ref{fig:our_embedding_space}, particularly with shorter sentences. Coupled with the improved performance on STS tasks, we can conclude that our framework is less prone to learning from spurious correlations.



\paragraph{Learned similarity metric.}
Returning to the example initially posed in \S\ref{sec:method}, we show in Figure \ref{fig:retrieval} similarity scores for pairs of examples computed by our BERT\ba model, as well as the corresponding DiffCSE and SimCSE variants. Notice that all three assign higher similarities between anchor: "A man is lifting weights in a garage", and phrases: "A man is lifting weights", "A man in a garage". However, despite their equal constitution in the anchor text, SimCSE incorrectly assesses a higher similarity between the anchor and the first phrase, whereas DiffCSE and our model better capture the equivalence in similarity.  The same occurs with anchor: "We store it outside of the house", and texts: "A man is in a garage", "She parked on the driveway"; despite both being unrelated to the anchor, SimCSE spuriously assigns a higher affinity to the former. Overall, we observed parity in the similarity assessments given by our model and DiffCSE, which validates the ability of our approach to remedy the suboptimal alignment of SimCSE without explicit incentive.


\section{Conclusion} In summary, we proposed a new way to construct positive pairs for unsupervised contrastive learning frameworks relying on pre-trained language models. Our experiments on STS tasks verified the effectiveness of the approach, which achieved competitive results with more complex learning methods, with the benefit of stabilizing and reducing the overall cost of training. We provided empirical studies and qualitative examinations into our approach, verifying its ability to train sentence encoders with better alignment. We believe this work can foster new avenues of inquiry in contrastive learning, especially those that draw upon a \textit{human} cognition of language.  

\newpage

\section{Limitations}

There are several limitations in this work. First, we have not explored how to make use of composition-based augmentations in the supervised setting. A second limitation is a lack of theoretical grounding in the impact of our latent space composition. Finally, we have not explored interoperability with other training objectives. 

\section{Note on Ethics}
\label{sec:ethics}

We do not believe there are significant ethical considerations stemming from our work, except those that accompany the use of language models and unlabelled corpora in general. Pre-trained language models, including BERT and RoBERTa, are known to learn and reiterate harmful prejudices. Although our pre-training corpus is sourced from Wikipedia and cited in several related works, it cannot be feasibly vetted for explicit or inappropriate content.

\section*{Acknowledgements}
We thank the anonymous reviewers for their valuable feedback and input. We also thank Haotian Xu, for his insight and suggestions that shaped the course of this work. We gratefully acknowledge support from National Science Foundation (NSF) via the awards IIS-1942918 and IIS-2127746. Portions of this research were conducted with the advanced computing resources provided by Texas A\&M High Performance Research Computing. 

\bibliographystyle{acl_natbib}
\bibliography{anthology, custom}


\end{document}